\documentclass[runningheads]{llncs}
\usepackage{color}
\usepackage{caption}
\usepackage{amsmath,amssymb}
\usepackage{amsfonts}
\usepackage{verbatim}
\usepackage{enumerate}
\usepackage[pdftex]{graphicx}
\usepackage{multirow}
\usepackage{times}

\usepackage{soul}
\usepackage{xurl}
\usepackage{hyperref}
\usepackage{color}
\usepackage{booktabs}
\usepackage{array}
\usepackage{bm}
\usepackage{breakcites}
\usepackage[title,toc,titletoc,page]{appendix}
\usepackage[misc,geometry]{ifsym}
\usepackage{url}
\usepackage[T1]{fontenc}
\usepackage{algorithm,algpseudocode}
\usepackage{amsmath,amssymb}
\usepackage{mathtools}

\usepackage{algpseudocode}
\usepackage[utf8]{inputenc}
\algrenewcommand\algorithmicindent{1.5em}
\algnewcommand{\Inputs}[1]{%
  \State \textbf{Inputs:} \hspace*{\algorithmicindent}\parbox[t]{.8\linewidth}{\raggedright #1}
}
\algnewcommand{\Outputs}[1]{%
  \State \textbf{Outputs:} \hspace*{\algorithmicindent}\parbox[t]{.8\linewidth}{\raggedright #1}
}
\algnewcommand{\Initialize}[1]{%
  \State \textbf{Initialize:}
  \Statex \hspace*{\algorithmicindent}\parbox[t]{1.2\linewidth}{\raggedright #1}
}
\usepackage[subrefformat=parens]{subcaption}
\begin{document}
\title{Meta-learning of Pooling Layers \\for Character Recognition\thanks{We provide our implementation at \url{https://github.com/Otsuzuki/Meta-learning-of-Pooling-Layers-for-Character-Recognition}.}}
\renewcommand\footnotemark{}
\titlerunning{Meta-learning of Pooling Layers for Character Recognition}

\author{Takato Otsuzuki\inst{1} \and
Heon Song\inst{1} \and
Seiichi Uchida\inst{1}\orcidID{0000-0001-8592-7566} \and \\
Hideaki Hayashi\inst{1} (\Letter) \orcidID{0000-0002-4800-1761}}
\authorrunning{T. Otsuzuki et al.}

\institute{Kyushu University, Fukuoka, Japan, \\
\email{takato.otsuzuki@human.ait.kyushu-u.ac.jp, heon.song@human.ait.kyushu-u.ac.jp, uchida@ait.kyushu-u.ac.jp, hayashi@ait.kyushu-u.ac.jp}}
\maketitle              
\begin{abstract}
In convolutional neural network-based character recognition, pooling layers play an important role in dimensionality reduction and deformation compensation. However, their kernel shapes and pooling operations are empirically predetermined; typically, a fixed-size square kernel shape and max pooling operation are used. In this paper, we propose a meta-learning framework for pooling layers. As part of our framework, a parameterized pooling layer is proposed in which the kernel shape and pooling operation are trainable using two parameters, thereby allowing flexible pooling of the input data. We also propose a meta-learning algorithm for the parameterized pooling layer, which allows us to acquire a suitable pooling layer across multiple tasks. In the experiment, we applied the proposed meta-learning framework to character recognition tasks. The results demonstrate that a pooling layer that is suitable across character recognition tasks was obtained via meta-learning, and the obtained pooling layer improved the performance of the model in both few-shot character recognition and noisy image recognition tasks. 

\end{abstract}

\section{Introduction} \label{sec:intro}
\begin{figure}[!t]
    \centering
    \includegraphics[width=1.0\linewidth]{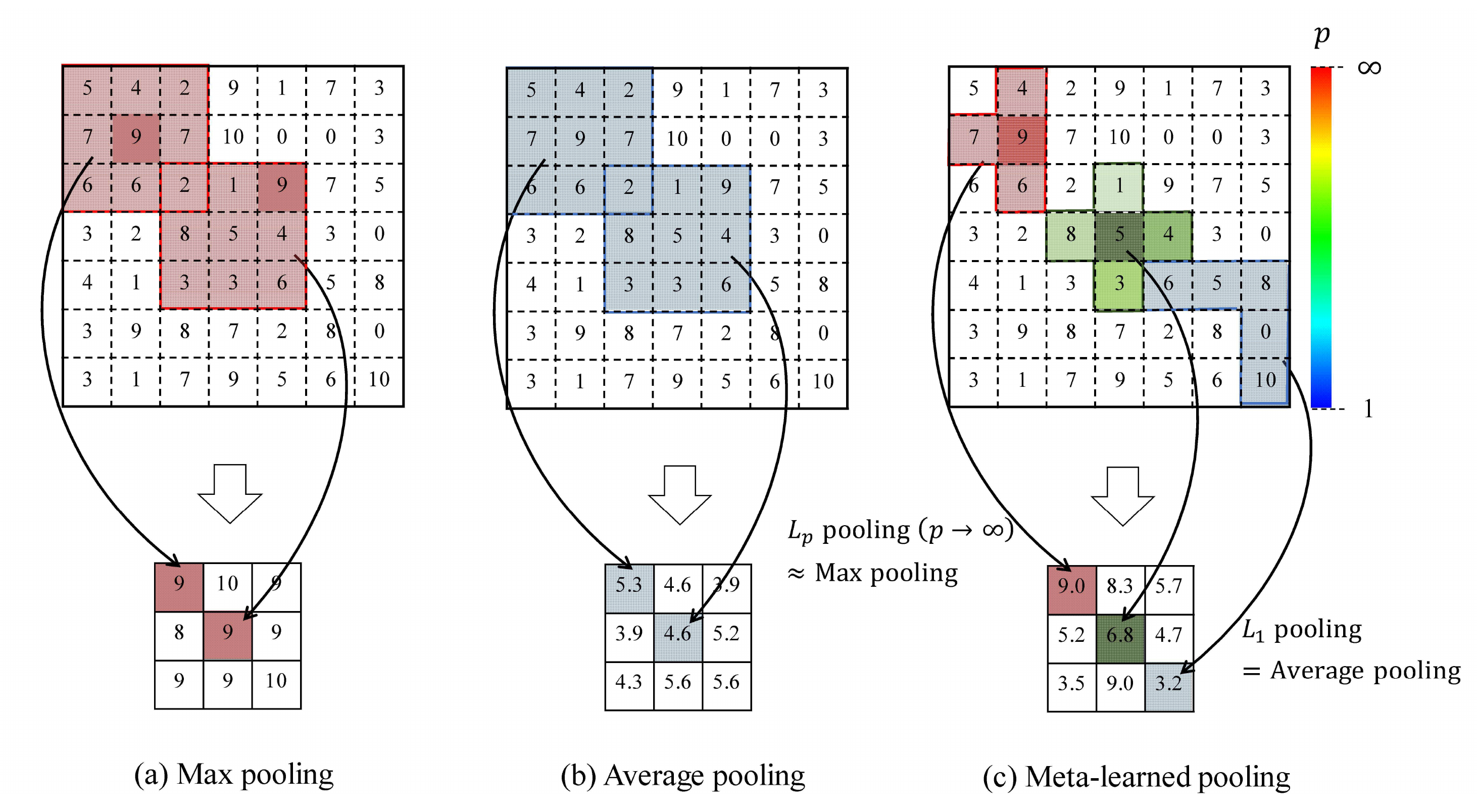}
    \caption{Meta-learning of pooling and comparison with other pooling methods: (a)~max pooling, (b)~average pooling, and (c)~meta-learned pooling (ours). The matrices in the top and bottom rows represent feature maps before and after pooling processing, respectively, where the number in each cell corresponds to the feature value in each pixel. The colored squares represent kernels, and their colors correspond to the type of operations. In (a) and (b), the maximum and average values are calculated for each kernel. Moreover, all kernels share the same shape and operation. In (c), the proposed meta-learned pooling employs different shapes and operations in individual kernels.}
    \label{fig:task}
\end{figure}

In convolutional neural network (CNN)-based character recognition, pooling layers play an important role in dimensionality reduction and deformation compensation. In particular, the max and average pooling layers, as illustrated in Figs.~\ref{fig:task}(a) and (b), respectively, are widely used in CNNs. These pooling operations are effective in absorbing the deformations that occur in character images. Even if the convolutional features undergo local changes owing to the deformation of the character image, the dimensionally-reduced feature map is invariant to such changes. As a result, a CNN with pooling layers is robust to character image deformation.\par

Nevertheless, typical pooling layers have a limitation in that the kernel shape and pooling operation should be determined empirically and manually. In the pooling layers, pooling operations are performed by sliding a fixed-size kernel, and the kernel size is predetermined as a hyperparameter. In two-dimensional pooling, the shape of the kernel is fixed to a square generally. Although various pooling operations have been proposed~\cite{Malinowski2013LearningSP,graham2014fractional,saeedan2018detail,Wei_2019_CVPR,otsuzuki2020regularized}, the type of pooling operation is chosen manually, and the same operation is performed in the layer.\par 

The purpose of this study is to determine a pooling layer suitable for character recognition tasks in a data-driven approach. Although max and average pooling layers with fixed-size square kernels are generally used in tasks using character images, it is unknown whether this is in fact appropriate. If we can obtain a suitable pooling layer across multiple tasks using character images, we can apply it to subsequent new tasks using character images, thereby leading to improvements in recognition accuracy.\par

For this purpose, we utilize meta-learning. Meta-learning aims to improve the learning performance of a machine learning model by providing datasets of multiple tasks and then acquiring knowledge shared among the tasks. Specifically, hyperparameters, such as a learning rate and initial weights, that are suitable across all the tasks are obtained via meta-learning. As a result, the meta-learned model can improve its learning efficiency when a new task is given. \par

In this paper, we propose a meta-learning framework for pooling layers. As part of the framework, a parameterized pooling layer is proposed in which the kernel shape and pooling operation can be trained using two parameters, thereby allowing flexible pooling of the data. We also propose a meta-learning algorithm for the parameterized pooling layer, which allows us to acquire a suitable pooling layer across multiple tasks.\par 

Fig.~\ref{fig:task} illustrates an example of a pooling layer obtained under our framework, alongside its comparison with max and average pooling layers. In the traditional max and average pooling layers shown in Figs.~\ref{fig:task}(a) and (b), respectively, fixed-size square kernels are used, and the same operation is performed in each kernel. In contrast, in the pooling layer meta-learned in our framework, flexible kernel shapes are used. Furthermore, the operation in each kernel is defined based on $L_p$ pooling~\cite{sermanet2012convolutional}, which means various types of operations, including max and average pooling, can be realized depending on the value of $p$.\par

\begin{figure}[!t]
    \centering
    \includegraphics[width=1.0\linewidth]{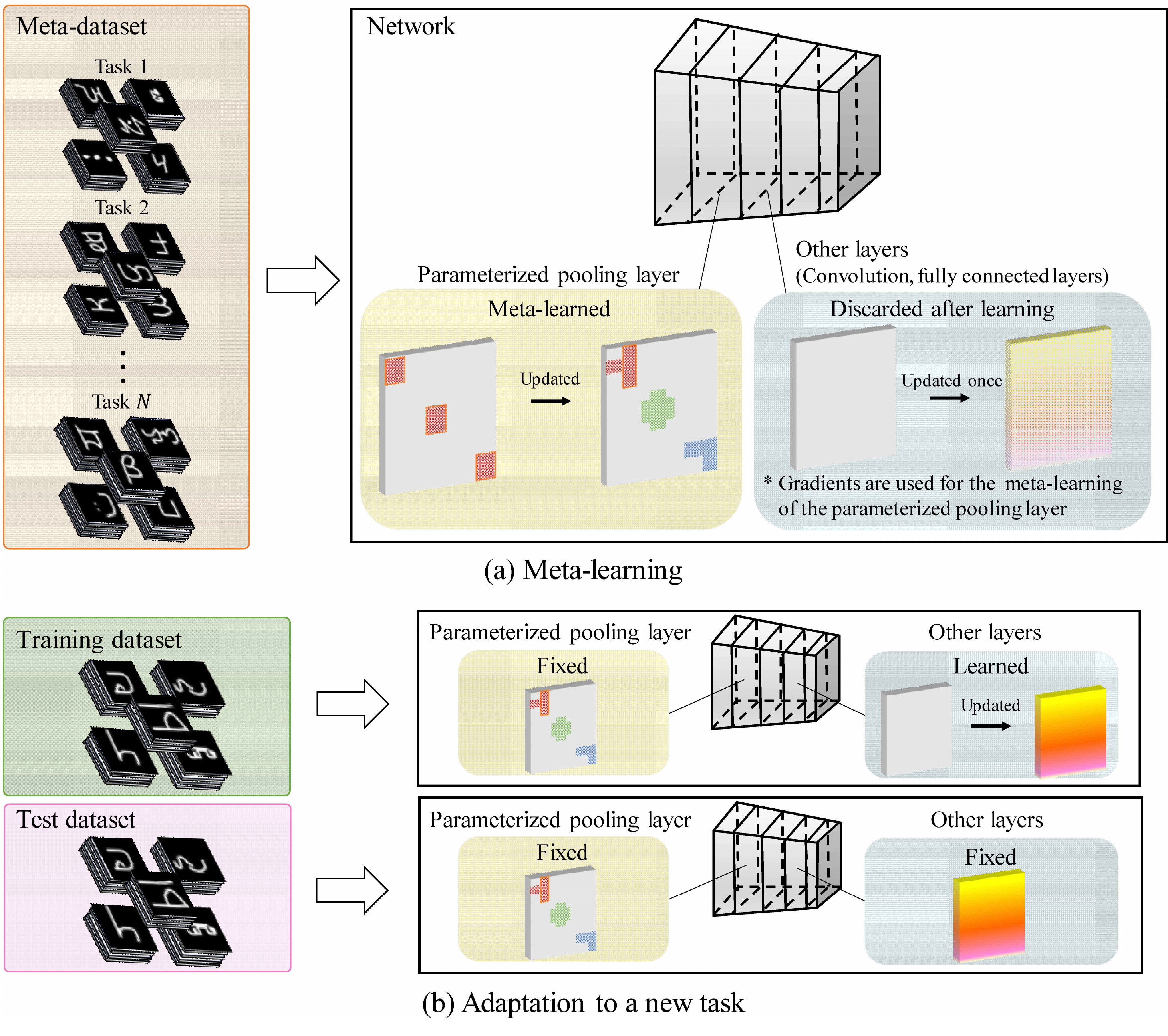}
    \caption{Meta-learning flow. The flow of the proposed framework is divided into two steps: (a) First, the kernel shapes and pooling operations of the parameterized pooling layer are trained via meta-learning. By learning the pooling layer from a large number of tasks, we can acquire knowledge across all tasks. At this time, the parameters of the other layers are updated once for each task batch, and their gradients are used in the meta-learning of the parameterized pooling layer, whereas the updates of the other layers themselves are not used in the subsequent step. (b) Then, the meta-learned pooling layer is adapted to a new task. In this step, with the meta-learned pooling fixed, the parameters of the remaining layers are trained using the training dataset of the new task. Next, with all parameters fixed, the trained model is adapted to the new task by classifying the test data.}
    \label{fig:overall}
\end{figure}

In the experiments, we reveal \textit{what pooling layer is suitable for character recognition tasks} in a data-driven approach and demonstrate that the obtained pooling layer improves the learning performance of a CNN for a subsequent new task. Fig.~\ref{fig:overall} shows the flow of the proposed method applied to the meta-learning of character image recognition. First, we prepare a meta-dataset containing a large number of character recognition tasks and apply the proposed meta-learning framework to a pooling layer in a CNN, as shown Fig.~\ref{fig:overall}(a). In this step, a pooling layer suitable for multiple character recognition tasks can be obtained. Then, the CNN with the meta-learned pooling layer is adapted to a new task (Fig.~\ref{fig:overall}(b)).\par

The main contributions of this study can be summarized as follows:
\begin{itemize}
    \item We propose a meta-learning framework for pooling layers. As part of the proposed framework, a parameterized pooling layer is proposed to make the kernel shape and pooling operation trainable. A meta-learning algorithm for the parameterized pooling layer is also presented.
    \item We reveal a pooling layer suitable for character recognition tasks in a data-driven manner. We prepare a large number of character image recognition tasks using the Omniglot dataset~\cite{lake2019omniglot} and then apply the proposed meta-learning framework to these tasks. After meta-learning, we analyze the obtained kernel shapes and pooling operations by visualizing the meta-parameters of the parameterized pooling layer. The results demonstrate that the meta-learned pooling layer applies max pooling around the character region in an annular shape. 
    \item We demonstrate the effectiveness of the meta-learned pooling layer for few-shot character image recognition and noisy character image recognition tasks. The meta-learned pooling layer is effective in improving accuracy for one-shot character image recognition compared to max and average pooling layers. The robustness to noisy images is also remarkably improved by the proposed method.
\end{itemize}


\section{Related Work}\label{sec:related}
\subsection{Pooling Layers}
In recent years, many pooling methods have been proposed and have demonstrated remarkable performance in specific tasks. Gao~\textit{et al.}~\cite{gao2019global} proposed a global second-order pooling (GSoP) block which can be used to calculate the second-order information in the middle layer of a CNN. Although methods that utilize the mean and variance of feature maps in the last layer have been proposed~\cite{lin2015bilinear,gao2016compact,Cui_2017_CVPR,wang2018multi,li2018towards}, the GSoP block is the first method that can be plugged in at any location of a CNN. Van~\textit{et al.}~\cite{nguyenvan2019pooling} proposed a novel scene text proposal technique, which iteratively applies a pooling operation to the edge feature extracted from an image, thereby improving the accuracy of scene text box detection. Hou~\textit{et al.}~\cite{Hou_2020_CVPR} proposed a strip pooling module, which considers a long but narrow kernel. They demonstrated that short-range and long-range dependencies among different locations can be captured simultaneously by combining a strip pooling module with a pyramid pooling module (PPM)~\cite{zhao2017pyramid} in semantic segmentation tasks.\par

Instead of utilizing a specific pooling operation, some trials have been aimed to generalize pooling operations for global use. A representative example is $L_p$ pooling~\cite{sermanet2012convolutional,feng2011geometric}, which calculates an $L_p$-norm over the input elements. Consequently, $L_p$ pooling involves average pooling when $p=1$ and max pooling when $p \rightarrow \infty$, and it can perform pooling operations between average and max pooling using $1 < p < \infty$ without completely eliminating the influence of input elements with small values. Another example is generalized mean pooling (GeM)~\cite{dollar2009integral,tolias2016particular,berman2019multigrain}, which is a generalization of average pooling and consequently has a similar operation to $L_p$ pooling.  A pooling method that can automatically enhance important features by learning has also been proposed~\cite{yu2014mixed,Gao_2019_ICCV}.\par 

The proposed method is inspired by the attempts to generalize pooling operations. In particular, the formulation of the proposed method is an expansion of $L_p$ pooling. The proposed method differs from $L_p$ pooling in that the kernel shape is trainable. Another difference from conventional trainable pooling methods is that the proposed method learns suitable pooling across multiple tasks, instead of a single task, based on meta-learning.

\subsection{Meta-learning}
Meta-learning aims to learn knowledge shared across multiple tasks to adapt the model to another subsequent task given with a small amount of training data or few learning steps. According to Baik~\textit{et al.}~\cite{Baik_2020_CVPR}, meta-learning algorithms can be divided into three categories: network-based, metric-based, and optimization-based. Network-based methods learn a fast adaptation strategy using auxiliary networks~\cite{pmlr-v48-santoro16,munkhdalai2017meta}. Metric-based methods learn the relationships between inputs in the task spaces and acquire object metrics or distance functions to facilitate problem solving~\cite{chen2020variational,NIPS2016_90e13578,sung2018learning}. Optimization-based methods adjust the optimization algorithm so that the model can be adapted to several tasks~\cite{finn2017model,Baik_2020_CVPR,rusu2018meta,zhou2021metalearning}.\par

Among existing optimization-based meta-learning algorithms, model-agnostic meta-learning (MAML)~\cite{finn2017model} has attracted much attention because of its simplicity and generality. MAML encodes prior knowledge so that the model can perform well across tasks and learn new tasks rapidly. Some studies have aimed to improve the performance of MAML while maintaining simplicity and generality. Rusu~\textit{et al.}~\cite{rusu2018meta} proposed an approach that learns an embedding of the model parameters into a low-dimensional space and conducts meta-learning in that space, achieving more efficient adaptation. Baik~\textit{et al.}~\cite{Baik_2020_CVPR} improved the MAML-based framework by attenuating the conflicts among tasks and layers, thereby improving its performance in few-shot learning. Some researchers have attempted to extend MAML. Zhou~\textit{et al.}~\cite{zhou2021metalearning} introduced a method called meta-learning symmetries by reparameterization (MSR) for meta-learning neural network architectures. They showed that MSR can learn symmetries shared among tasks from data based on meta-learning and reparameterization of network weights.\par

Although the algorithm of the proposed method is inspired by MAML and MSR, the proposed algorithm differs from the above optimization-based meta-learning algorithms in that meta-learning is applied to the pooling layer while considering the gradients of other layers. To the best of our knowledge, this is the first attempt in adopting meta-learning in the pooling layer. Another unique feature of the proposed algorithm is that the kernel shape matrix is learned with a constraint to be binary.

\section{Meta-learning of Pooling Layers}\label{sec:method}
In this section, we present our meta-learning framework for pooling layers. To make pooling layers trainable, we first introduce a parameterized pooling layer, wherein the shape and operation of each kernel are trained individually. We then describe the meta-learning algorithm of the parameterized pooling layer.\par 

\subsection{Parameterized Pooling Layer}\label{section3_1}
The purpose of the proposed method is to acquire the knowledge shared across multiple tasks, thereby developing a pooling layer that can be efficiently applied to new tasks. By parameterizing the pooling layer and making it trainable, flexible determination of kernel shapes and pooling operations can be achieved.\par 

The parameterized pooling layer has two (meta-)parameters that are trained via meta-learning. One is a kernel shape matrix, $\bm{W}$, with binary elements that determines the area in which the pooling operation is applied. By multiplying $\bm{W}$ with the input vector/image, various types of kernel locations and shapes can be realized, including the sliding square kernels of ordinary pooling. The other is an operation parameter, $\bm{p}$. The parameterized pooling layer is based on $L_p$ pooling, where the pooling operation in each kernel is defined by the $L_p$ norm. Incidentally, $L_p$ pooling is a generalization of pooling operations that includes max pooling (when $p \rightarrow \infty$) and average pooling (when $p = 1$). Whereas the value of $p$ in $L_p$ pooling is determined empirically, the value of $\bm{p}$ in the proposed method is automatically determined via meta-learning.\par 

Given an input vector $\bm{x} \in \mathbb{R}^J_+$, the parameterized pooling layer $\bm{f}: \mathbb{R}_+^J \mapsto \mathbb{R}^I$ ($J \geq I$) is defined as follows:
\begin{equation}
\label{eq:param_pool}
f_i(\bm{x}) = \left(\frac{1}{J}\sum_{j=1}^J{W_{ij}x_j^{p_i}}\right)^\frac{1}{p_i}, \quad i = 1, \ldots I, 
\end{equation}
where $\bm{W} \in \{0, 1\}^{I \times J}$ is the kernel shape matrix, $\bm{p} \in \mathbb{R}_+^I$ is the operation parameter, and $J$ and $I$ are the input and output dimensions, respectively. Although the assumption that the input elements are positive is relatively strong, in practice, it can be satisfied by applying a rectified linear unit (ReLU) activation function to the previous layer.\par

The parameterized pooling layer can also be applied to an image input, as with conventional pooling layers. One simple method is to vectorize the image input by raster scanning before inputting the data to the parameterized pooling layer. Another method is to combine equation \eqref{eq:param_pool} and sliding windows. Taking a small window in the input image, we vectorize the pixels in the window and prepare independent $\bm{W}$ and $\bm{p}$ for each window. Then, we apply equation \eqref{eq:param_pool} to the entire image while shifting the window across the image. In practice, the latter method is computationally advantageous; therefore, we employed it in our experiments using image inputs, as reported in Sections \ref{sec:simulation} and \ref{sec:character_rec}.\par

To maintain the binariness of $\bm{W}$ and positiveness of $\bm{p}$, we employ the following variable transformation: 
\begin{align}
    W_{ij} &= \begin{cases}
        \mathrm{Sigmoid}((\tilde{W}_{ij} - 0.5)/T) & \text{during meta-learning}  \\
        \mathrm{Step}(\tilde{W}_{ij} - 0.5) & \text{otherwise}
    \end{cases}, \\
    p_i &= \exp{(\tilde{p}_i)},
\end{align}
where $\mathrm{Sigmoid}(\cdot)$ is the sigmoid function, $T$ is a temperature parameter, $\mathrm{Step}(\cdot)$ is the unit step function, which returns 0 for a negative input and 1 otherwise, and $\tilde{W}_{ij} \in \mathbb{R}$ and $\tilde{p}_i \in \mathbb{R}$ are auxiliary variables. The step function is used to make $\bm{W}$ binary, but it is not differentiable; therefore, the sigmoid function is used as an approximation during meta-learning to enable gradient calculation. For $\bm{p}$, the exponential function is used to satisfy $p_i > 0$. In the meta-learning algorithm described below, the gradients are calculated with respect to the auxiliary variables $\tilde{W}_{ij}$ and $\tilde{p}_i$. However, in the following sections, we will explain as if we deal directly with $W_{ij}$ and $p_i$ instead of $\tilde{W}_{ij}$ and $\tilde{p}_i$ for simplicity.

\subsection{Meta-learning Algorithm}\label{section3_2}
The aim of the meta-learning of pooling is to learn and exploit a pooling layer that is suitable for multiple tasks. Given a task distribution $\rho\bigl(\tau\bigl)$, we assume that there is an optimal pooling layer that is shared across $\rho\bigl(\tau\bigl)$, and we estimate it by learning meta-parameters $\bm{W}$ and $\bm{p}$ through meta-learning.\par 

\begin{algorithm}[t]
\caption{Meta-learning of the parameterized pooling layer}
\label{alg:main}
\begin{algorithmic}[0]
\Inputs{$\{\tau_j\}^N_{j=1} \sim \rho\bigl(\tau\bigl)$: Meta-learning tasks}
\Inputs{\{$\bm{W}$, $\bm{p}$\}: Randomly or arbitrarily initialized meta-parameters}
\Inputs{$\bm{\theta}$: Randomly initialized weights of other layers}
\Inputs{$\alpha$, $\eta$: Learning rates for the inner and outer loops}
\While {$not \ done$}
\State{Sample minibatch $\{\tau'_i\}^n_{i=1} \sim \{\tau_j\}^N_{j=1}$}
\For{$\tau'_i \in \{\tau'_i\}^n_{i=1}$}
  \State $\{D^\mathrm{tr}_{i}, D^\mathrm{val}_{i}\} \leftarrow \ \tau'_{i}$; \ // task data
  \State $\bm{\delta}_{i} \leftarrow \ \nabla_{\bm{\theta}}\mathcal{L}\bigl(\bm{W}, \bm{p}, \bm{\theta},  D^\mathrm{tr}_i\bigl)$;
  \State $\bm{\theta}^{\prime} \leftarrow \ \bm{\theta} - \alpha\bm{\delta}_i$; \ // inner step
  \State /* outer gradient */
  \State $\bm{G}_{i}^\mathrm{w} \leftarrow \frac{\mathrm{d}}{\mathrm{d}\bm{W}}\mathcal{L}\bigl(\bm{W},\bm{p},\bm{\theta}^{\prime},D^\mathrm{val}_{i}\bigl)$, $\bm{G}_{i}^\mathrm{p} \leftarrow \nabla_{\bm{p}}\mathcal{L}\bigl(\bm{W},\bm{p},\bm{\theta}^{\prime},D^\mathrm{val}_{i}\bigl)$;
\EndFor
\State $\bm{W} \leftarrow \bm{W} - \eta\sum_{i} \bm{G}_{i}^\mathrm{w}$, \ $\bm{p} \leftarrow \bm{p} - \eta\sum_{i} \bm{G}_{i}^\mathrm{p};$
\EndWhile
\end{algorithmic}
\end{algorithm}
Algorithm~\ref{alg:main} shows the meta-learning algorithm of the parameterized pooling layer. The algorithm is a gradient-based meta-learning algorithm similar to MSR~\cite{zhou2021metalearning} and MAML~\cite{finn2017model} algorithms. Assuming that we have a neural network with a parameterized pooling layer as well as other layers with weights $\bm{\theta}$, the meta-parameters $\bm{W}$ and $\bm{p}$ are updated during the meta-learning algorithm. Note that even if there are multiple parameterized pooling layers in a neural network, we can independently apply this algorithm to each one.

This algorithm consists of inner and outer loops. The inner loop calculates gradients and is nested in the outer loop, which updates the meta-parameters. In the outer loop, a task minibatch $\{\tau'_i\}^n_{i=1}$ is first sampled from the task set $\{\tau_j\}^N_{j=1}$, where $n$ and $N$ are the minibatch size and number of tasks, respectively. The task minibatch is then split into training and validation data $\{D^{\textrm{tr}}_{i}, D^{\textrm{val}}_{i}\}$ for any $\tau'_i$. Proceeding to the inner loop, the parameters of other layers $\bm{\theta}$ are updated as $\bm{\theta}^{\prime} \leftarrow \bm{\theta} - \alpha\nabla_{\bm{\theta}}\mathcal{L}\bigl(\bm{W}, \bm{p}, \bm{\theta},  D^\mathrm{tr}_i\bigl)$, where $\mathcal{L}(\cdot)$ is the loss function and $\alpha$ is the learning rate for the inner loop. After calculating the gradients with respect to the meta-parameters using the updated $\bm{\theta}^{\prime}$ and validation data, the meta-parameters are updated in the outer loop as follows:
\begin{align}
\label{eq:outer_grad}
\bm{W} &\leftarrow \bm{W} - \eta\sum_{i} \frac{\mathrm{d}}{\mathrm{d}\bm{W}}\mathcal{L}\bigl(\bm{W},\bm{p},\bm{\theta}^{\prime},D^\mathrm{val}_{i}\bigl), \\
\bm{p} &\leftarrow \bm{p} - \eta\sum_{i} \nabla_{\bm{p}}\mathcal{L}\bigl(\bm{W},\bm{p},\bm{\theta}^{\prime},D^\mathrm{val}_{i}\bigl),
\end{align}
where $\eta$ is the learning rate for the outer loop.

\section{Experiment on Artificial Data}\label{sec:simulation}
To verify the validity of the proposed method, we conducted an experiment using artificially generated data. The goal of this experiment is to verify whether the proposed method can meta-learn the kernel shapes and pooling operations from a task set that is generated by passing random vectors or images through a certain pooling layer.\par

\begin{figure}[!t]
    \centering
    \includegraphics[width=0.8\linewidth]{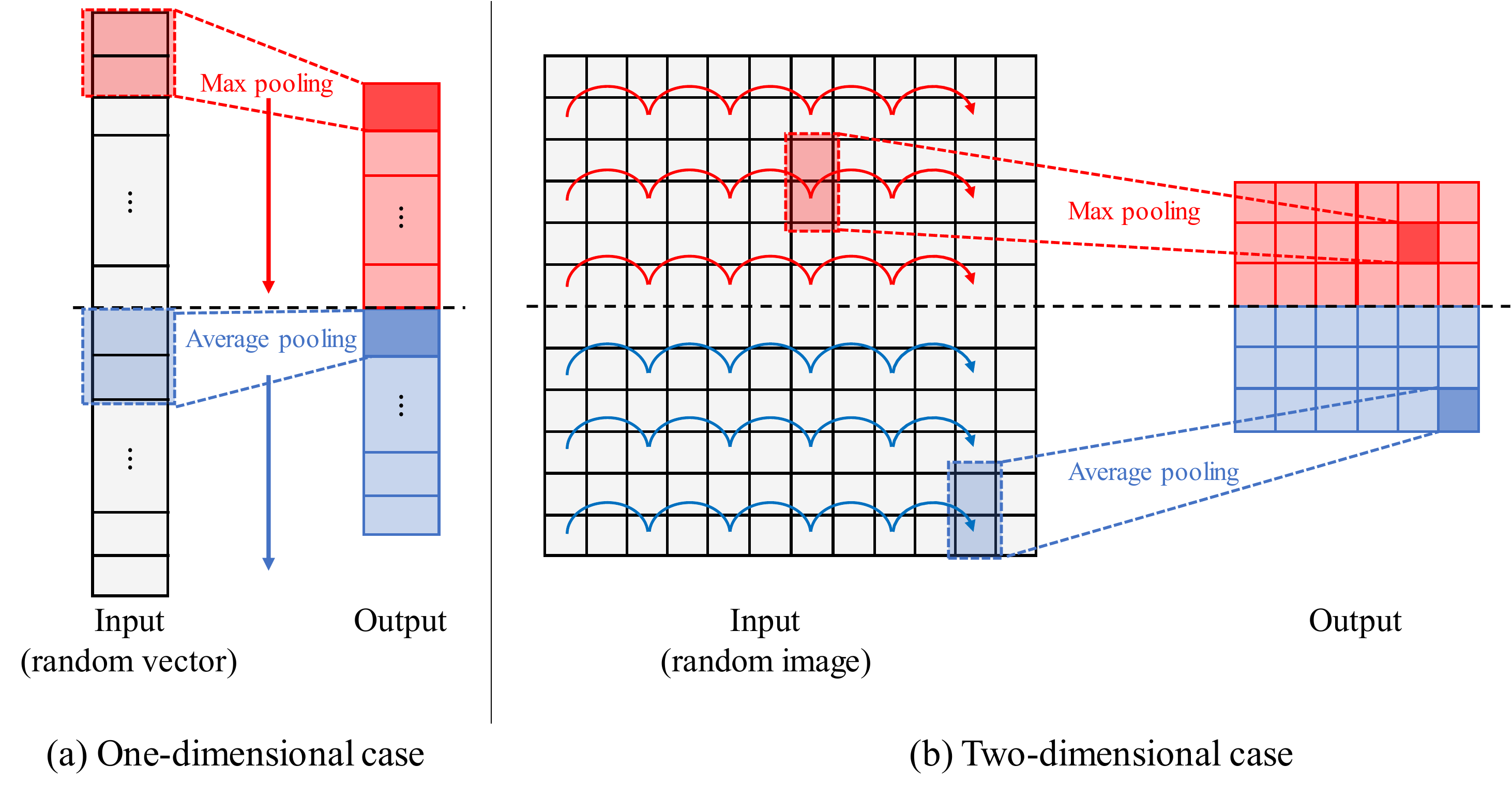}
    \caption{Data generation process for the experiment using artificial data. (a) In the one-dimensional case, a set of input and output data was generated by passing a random vector through a 1D pooling layer with a kernel size of 2 and stride of 2. (b) In the two-dimensional case, a random image was passed through a 2D pooling layer with a kernel size of $2 \times 1$ and stride of 2, i.e., pooling with a vertical rectangular kernel was performed for every other column. In both cases, max pooling was used in the first half, and average pooling was used in the second half.}
    \label{fig:simulation}
\end{figure}

Fig.~\ref{fig:simulation} outlines the data generation process for this experiment. We prepared datasets for one-dimensional and two-dimensional cases. For the one-dimensional case (Fig.~\allowbreak\ref{fig:simulation}(a)), we generated sets of inputs and outputs by passing random input vectors, whose elements were generated from a uniform distribution over [0, 1], through a 1D pooling layer with a filter size of $2$ and a stride of $2$. For the two-dimensional case (Fig.~\ref{fig:simulation}(b)), we generated sets of inputs and outputs by passing a random image, in which each pixel value was generated from a uniform distribution over [0, 1], through a 2D pooling layer with a filter size of $2 \times 1$ (vertical rectangle) and a stride of $2$. Here, the pooling layer consisted of max pooling for the first half and average pooling for the second half. Each task contains 20 sets of input and output vectors/images, and we generated $8{,}000$ training tasks. We set the dimensions of the input and output data to $60$ and $30$ in the one-dimensional case, and $28 \times 28$ and $14 \times 14$ in the two-dimensional case, respectively.\par

In the meta-learning, we used a neural network with only a single parameterized pooling layer. We set the task batch size to $32$ and the number of epochs in the outer loop to $10{,}000$, which was sufficient to converge the training. The data in each task were split into $1$ training sample and $19$ validation samples. In the outer loop, we used the Adam~\cite{kingma2014adam} optimizer with a learning rate of $0.001$. The temperature parameter $T$ was set to 0.2. In the inner loop, $\bm{W}$ and $\bm{p}$ themselves were updated only once instead of $\bm{\theta}$ because the network used in this experiment did not have any layer other than pooling, and the stochastic gradient descent (SGD) optimizer with a learning rate of $0.1$ was used. We used the mean squared error as the loss function. After meta-learning, we observed the learned meta-parameters $\bm{W}$ and $\bm{p}$. \par

\begin{figure}[t]
  \centering
  \begin{minipage}[b]{0.45\linewidth}
    \centering
    \includegraphics[keepaspectratio, width=1.0\hsize]{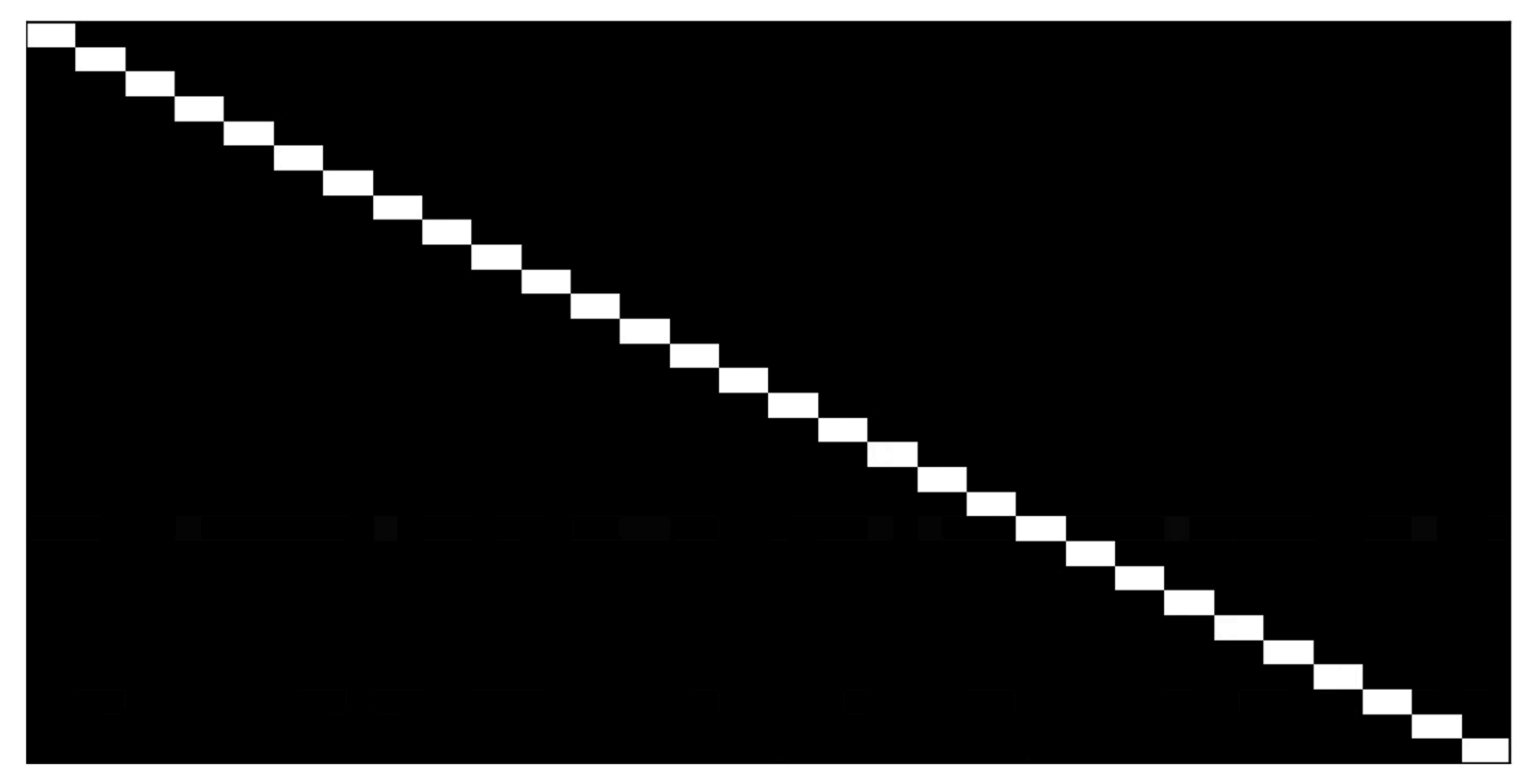}
    \subcaption{Kernel shape matrix $\bm{W}$}\label{fig:fig3_w}
  \end{minipage}
  \begin{minipage}[b]{0.45\linewidth}
    \centering
    \includegraphics[keepaspectratio, width=1.0\hsize]{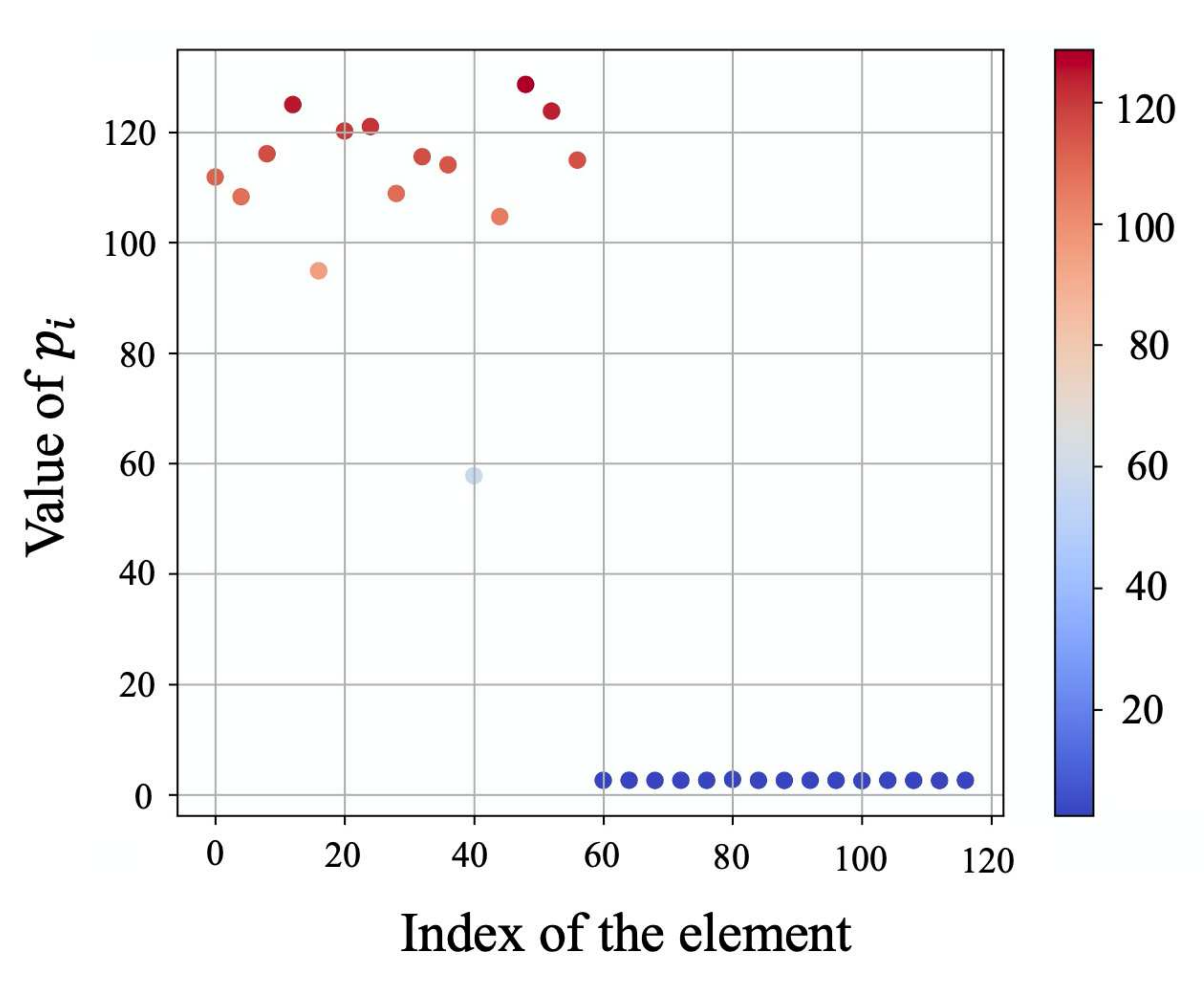}
    \subcaption{Operation parameter $\bm{p}$}\label{fig:fig3_p}
  \end{minipage}
  \caption{Learned meta-parameters $\bm{W}$ and $\bm{p}$ in the simulation experiment on one-dimensional artificial data. In (a), the black and white pixels represent 0 and 1, respectively. The numbers of columns and rows correspond to the input and output dimensions, respectively. }\label{fig:sim_param_1d}
\end{figure}

\begin{figure}[t]
  \centering
  \begin{minipage}[b]{0.35\linewidth}
    \centering
    \includegraphics[keepaspectratio, width=0.89\hsize]{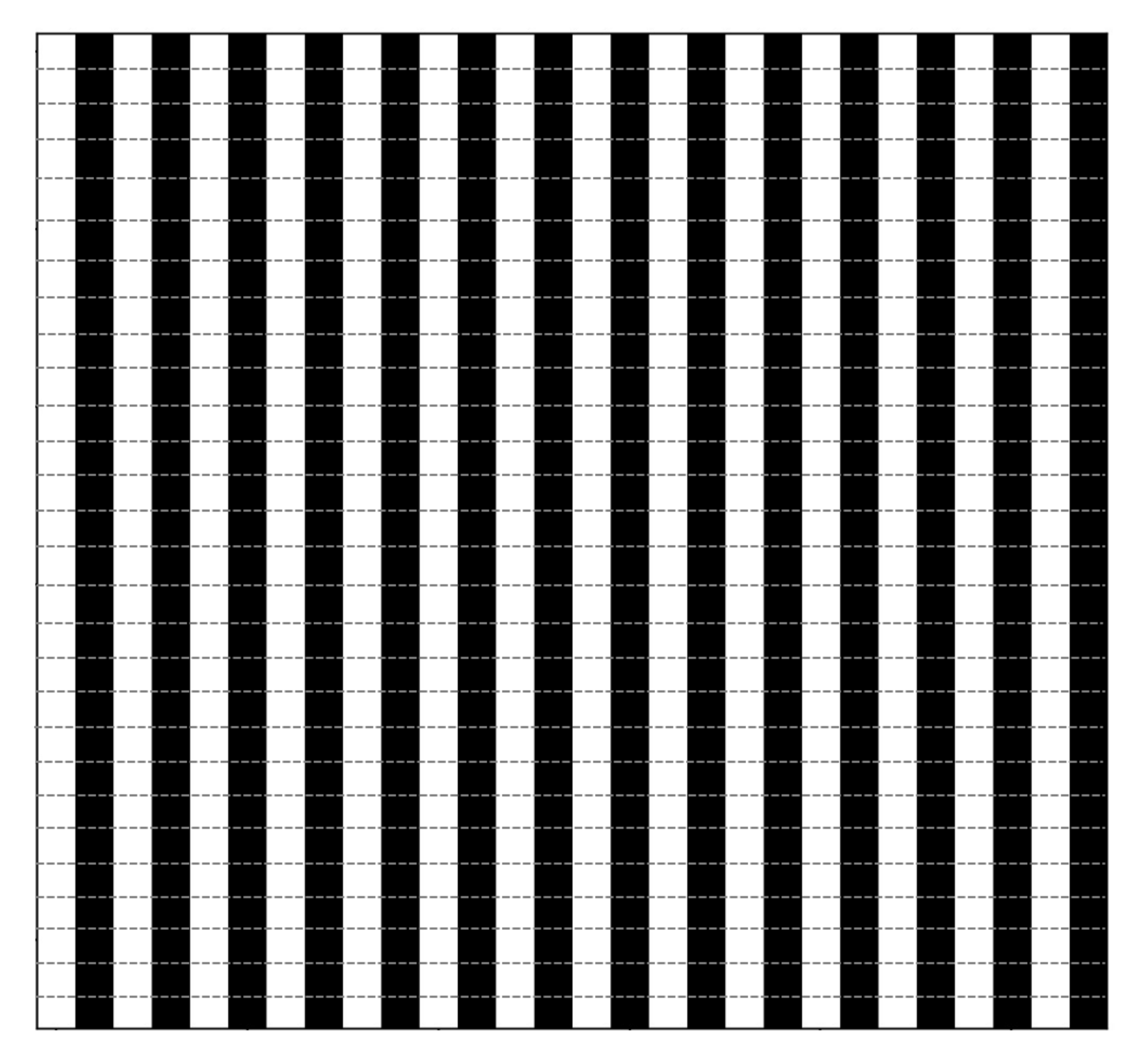}
    \subcaption{Kernel shape matrix $\bm{W}$}\label{fig:2d_w}
  \end{minipage}
  \begin{minipage}[b]{0.35\linewidth}
    \centering
    \includegraphics[keepaspectratio, width=1.0\hsize]{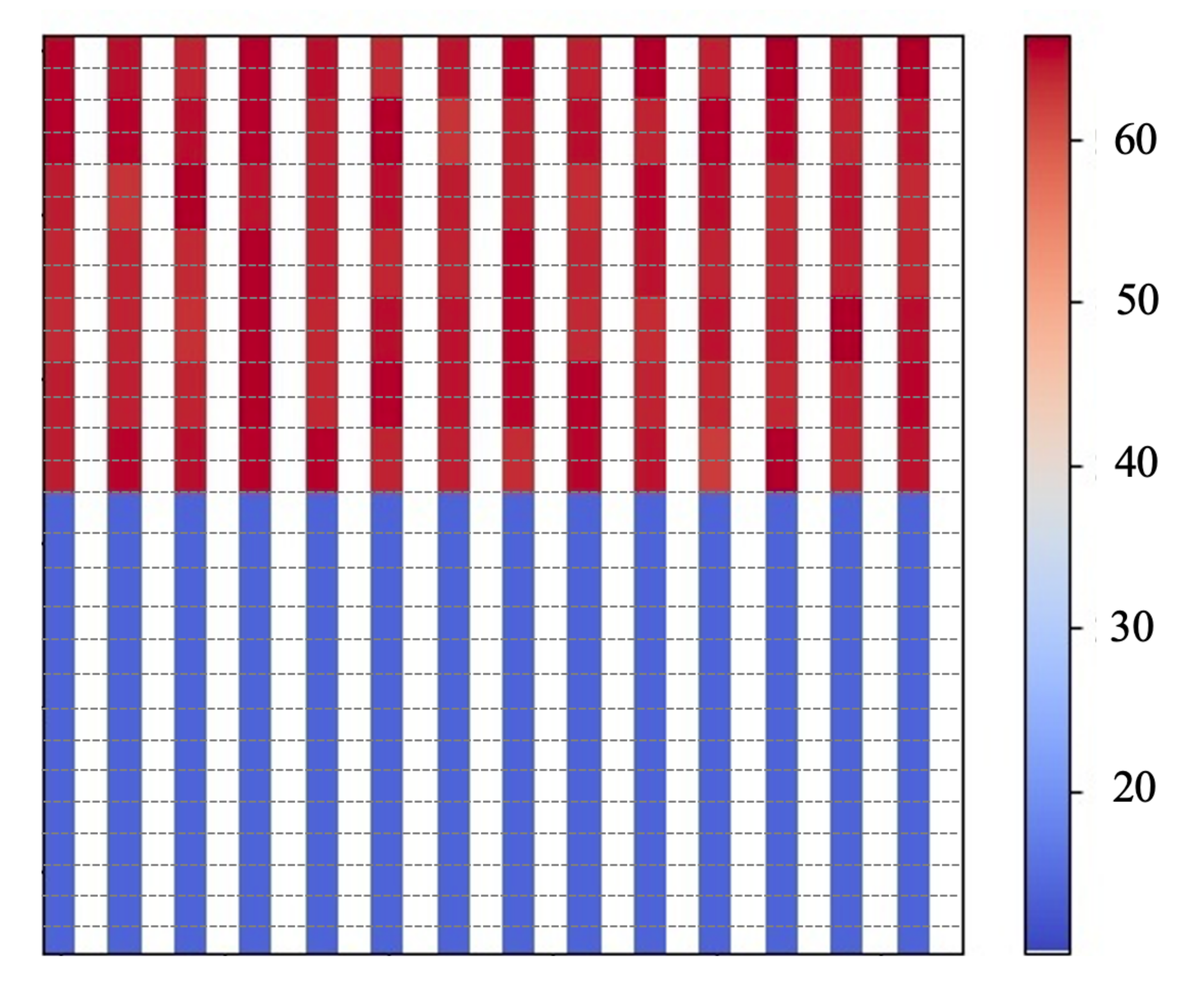}
    \subcaption{Operation parameter $\bm{p}$}\label{fig:2d_p}
  \end{minipage}
  \caption{Learned meta-parameters $\bm{W}$ and $\bm{p}$ in the simulation experiment on two-dimensional artificial data. In (a), the black and white pixels represent 0 and 1, respectively.}\label{fig:sim_param_2d}
\end{figure}

Figs.~\ref{fig:sim_param_1d} and~\ref{fig:sim_param_2d} show the learned meta-parameters $\bm{W}$ and $\bm{p}$ for the one-dimensional and two-dimensional artificial data, respectively. In Fig.~\ref{fig:sim_param_1d}(a), higher values are concentrated on the diagonal parts of $\bm{W}$, demonstrating that the parameterized pooling layer successfully learned the kernel shape with a size of 2 and stride of 2. It can also be confirmed that $\bm{p}$ takes large values in the first half and approximates $1$ in the second half, meaning that the max and average pooling operations are also learned approximately. In Fig.~\ref{fig:sim_param_2d}(a), the pooling pattern with a kernel size of $2 \times 1$ and stride of $2$ was reproduced. The pooling operations with max pooling in the first half and average pooling in the second half were also correctly estimated as shown in Fig.~\ref{fig:sim_param_2d}(b).\par

These results demonstrate that the proposed method can learn the kernel shapes and pooling operations in a data-driven manner. In this experiment, we used a network with only a parameterized pooling layer for simplicity. We have also confirmed that the parameterized pooling layer can be meta-learned in a network that includes other layers, such as convolutional and fully connected layers, but these results are omitted due to page limitations. In the next section, we describe experiments using character images with a network that includes other layers besides a parameterized pooling layer. 

\section{Experiment on Character Datasets}\label{sec:character_rec}
To evaluate the effectiveness of the proposed method on real-world datasets, we conducted a character image recognition experiment. The aim of this experiment is twofold: One is to determine a pooling layer suitable for character recognition tasks in a data-driven manner, and the other is to evaluate the generalization capability of the meta-learned pooling for new tasks. For the former, we first prepared a CNN that includes a parameterized pooling layer. Then, we meta-learned the parameterized pooling layer using multiple character recognition tasks. After that, we analyzed the learned meta-parameters of the pooling layer, which can be regarded as the kernel shapes and pooling operations that are suitable across multiple tasks. For the latter, we evaluated the model using tasks that were not given during the meta-training. Based on the assumption that the proposed meta-learning of pooling can obtain the knowledge shared across multiple datasets, meta-learned pooling will improve the generalization capability of the model for subsequent new tasks. Therefore, we evaluated the model with the meta-learned pooling layer based on few-shot image recognition and noisy image recognition tasks. Fixing the meta-parameters, we trained the weights of the remaining layers and evaluated the entire model.

\subsection{Dataset}
\begin{figure}[!t]
    \centering
    \includegraphics[width=0.55\linewidth]{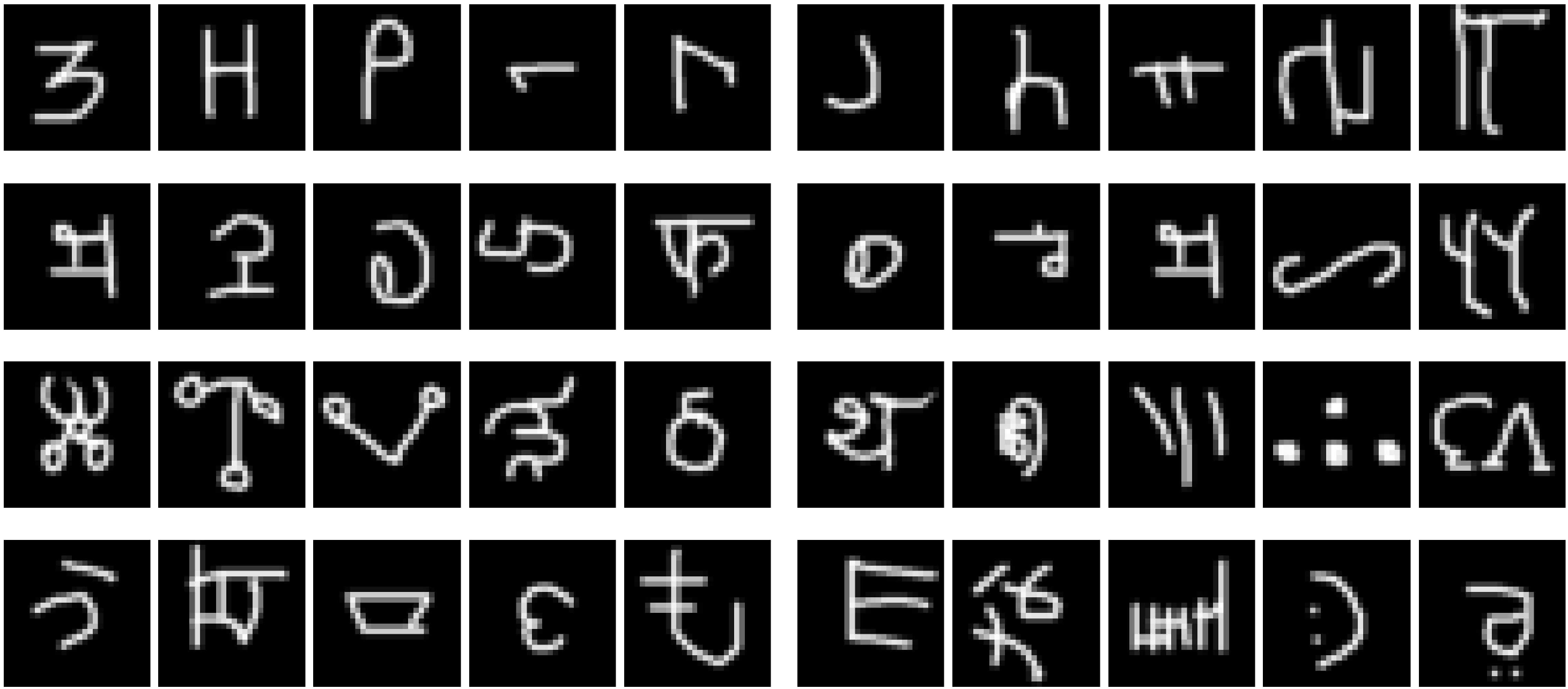}
    \caption{Example images from the Omniglot dataset. Pixel colors were inverted from the original images.}
    \label{fig:omniglot}
\end{figure}

We used the Omniglot dataset~\cite{lake2019omniglot}, which is a standard benchmark dataset of handwritten character images. Examples from the Omniglot dataset are shown in Fig.~\ref{fig:omniglot}. In particular, this dataset is frequently used in few-shot learning tasks~\cite{NIPS2016_90e13578,Elsken_2020_CVPR,khodadadeh2019unsupervised}. In this study, we used this dataset to meta-learn a suitable pooling layer for multiple character recognition tasks and evaluate the meta-learned pooling in few-shot and noisy image recognition tasks. The Omniglot dataset consists of $20$ instances of $1{,}623$ characters from $50$ different alphabets, and each instance was depicted by a different person. In this experiment, we randomly selected $1{,}200$ characters for meta-learning regardless of the alphabet, and used the remaining $423$ characters for few-shot and noisy image recognition tasks. We used the images after inverting pixel colors. We applied data augmentation by rotating the images by 90, 180, or 270 degrees.\par

\subsection{Experimental Setups}
Through the experiment, we used the same network architecture. The network consisted of a $3 \times 3$ convolutional layer with $64$ filters, batch normalization~\cite{ioffe2015batch}, ReLU activation functions, a $2 \times 2$ parameterized pooling layer, and a fully connected layer with a softmax activation function. In the parameterized pooling layer, sliding windows with a size of $2 \times 2$ and a stride of $2$, i.e., no overlap between windows, were used. The temperature parameter $T$ was set to 0.2.\par

After meta-learning, we conducted few-shot recognition and noisy image recognition tasks. In the few-shot recognition task, the model classified images into one of five classes (five-way recognition) given either one or five images as the training data (one-shot or five-shot, respectively). Among the $423$ character classes that were not used in the meta-learning, five classes were randomly sampled 100 times with duplication to generate 100 five-way recognition tasks. In each task, one or five images were used to train the weights of the convolutional and fully connected layers, and the remaining images were used as test data to evaluate the recognition accuracy. In the noisy image recognition task, we reused the network weights obtained in the few-shot recognition tasks and evaluated the recognition accuracy by adding salt-and-pepper noise to the test data. We varied the noise ratio in the interval of 10--60\%.\par 

In the meta-learning, the number of task batches was set to $32$, and the images in each batch were split into $1$ for training and $19$ for validation in the one-shot setting and $5$ for training and $15$ for validation in the five-shot setting. We used the Adam optimizer with a learning rate of $0.001$ for the outer loop and the SGD optimizer with a learning rate of $0.1$ for the inner loop. The numbers of epochs for inner and outer loops were set to $1$ and $10{,}000$, respectively. The initial weights of the convolutional and fully connected layers were determined based on the MAML algorithm~\cite{finn2017model} using the meta-learning dataset.\par 

For comparison, we prepared two networks with the parameterized pooling layer replaced with either a max or average pooling layer. The results of the few-shot recognition and noisy image recognition tasks were compared to those of the prepared networks. The initial weights of these networks were also determined based on the MAML algorithm.\par

\subsection{Results}
\begin{figure}[t]
  \centering
  \begin{minipage}[b]{0.35\linewidth}
    \centering
    \includegraphics[keepaspectratio, width=0.8\hsize]{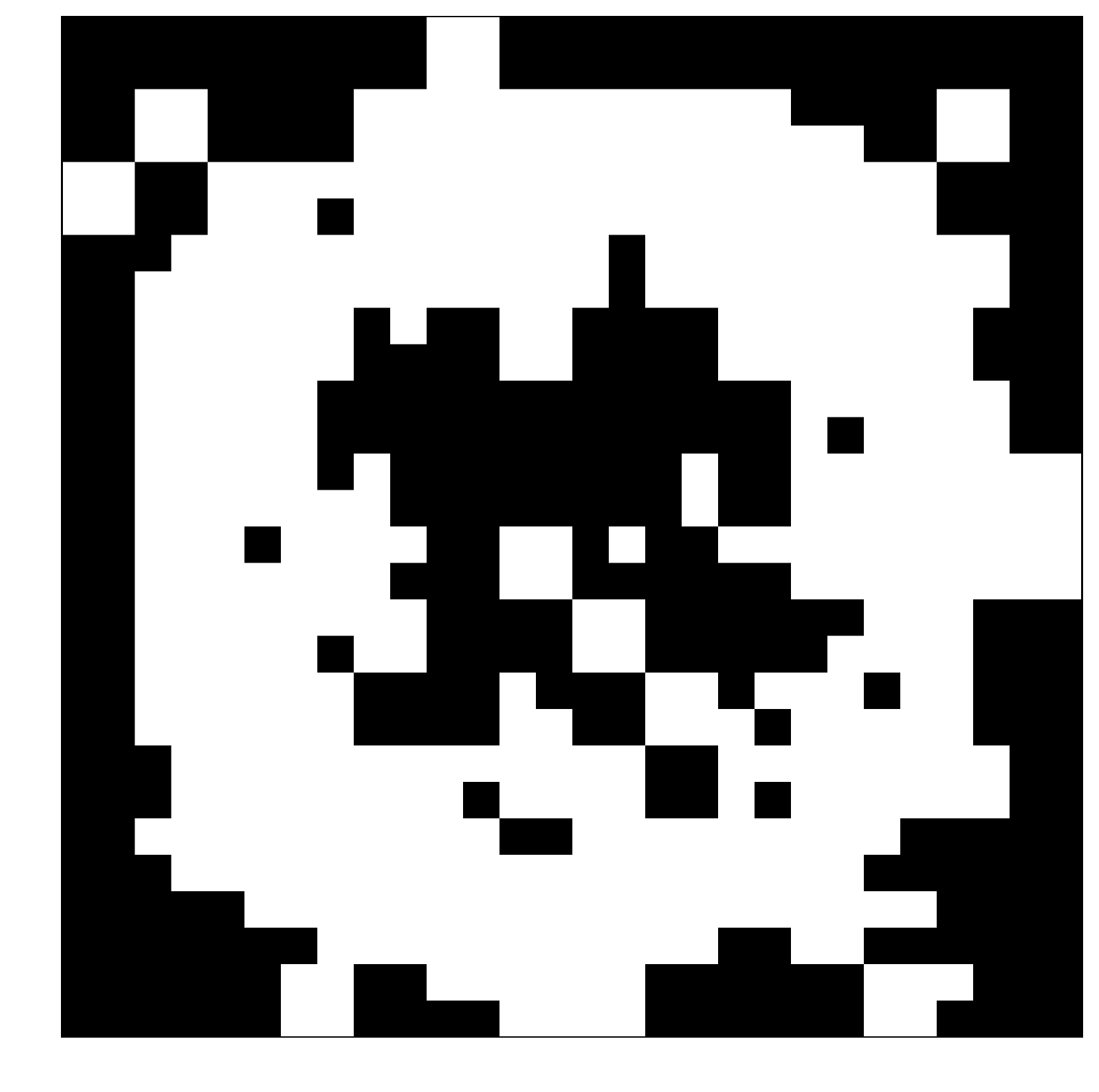}
    \subcaption{Kernel shape matrix $\bm{W}$}\label{fig:omniglot_w}
  \end{minipage}
  \begin{minipage}[b]{0.35\linewidth}
    \centering
    \includegraphics[keepaspectratio, width=1.0\hsize]{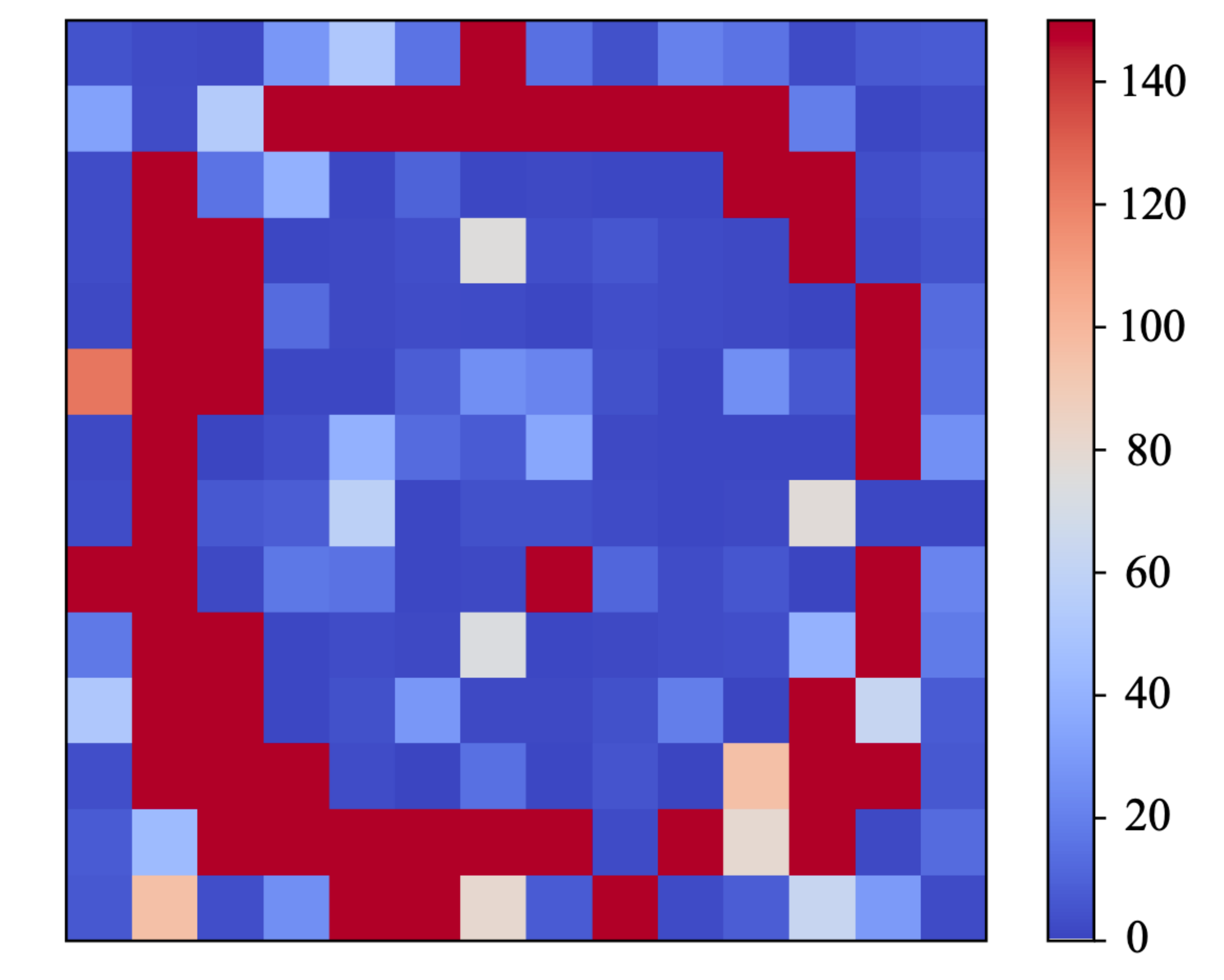}
    \subcaption{Operation parameter $\bm{p}$}\label{fig:omniglot_p}
  \end{minipage}
  \caption{Learned meta-parameters on the Omniglot dataset. (a) Kernel shape matrix $\bm{W}$ and (b) operation parameter $\bm{p}$. In (a), the black and white pixels represent 0 and 1, respectively. These meta-parameters were learned using non-overlapping sliding windows with a size of $2 \times 2$ and a stride of $2$. These figures were depicted by placing the learned parameters at the corresponding positions.}\label{fig:OmniglotParameter}
\end{figure}

The learned meta-parameters $\bm{W}$ and $\bm{p}$ are shown in Fig.~\ref{fig:OmniglotParameter}. According to the kernel shape matrix $\bm{W}$, the proposed pooling layer learned to extract the annular-shaped area around the region where the character is written. Surprisingly, it did not focus on the center of the image. From the result of $\bm{p}$, max pooling was applied around the character region, whereas average pooling was used in other regions. The annular shape observed in both $\bm{W}$ and $\bm{p}$ occurred because the character was not always written in the exact center of the image in the Omniglot dataset, and the essential information existed in somewhat peripheral regions. Furthermore, the white annulus in $\bm{W}$ was thicker than the red annulus in $\bm{p}$, indicating that max pooling was learned at the outer edge of the annulus to distinguish the character from the background, and average pooling was used in the area near the character to mitigate the effect of noise in the same manner as a linear filter.\par

Table~\ref{table:classification} lists the recognition accuracies for each pooling method in the one-shot and five-shot settings. As can be seen, the proposed meta-learned pooling showed the best performance in the one-shot setting, whereas max pooling outperformed ours in the five-shot setting. These results indicate that the proposed method is particularly effective when the training data are extremely limited.
\begin{table}[t]
\caption{Accuracies in the few-shot character recognition task.}
\label{table:classification}
\centering
\begin{tabular}{l||ccccccc}
\hline
                     & \multicolumn{2}{c}{Accuracy (\%)} \\
Pooling Operation    & One-shot              & Five-shot               \\ \hline
Max pooling          & $90.52 \pm 0.68$      & $\bm{97.3 \pm 0.92}$ \\
Average pooling      & $90.25 \pm 0.81$      & $97.0 \pm 0.93$      \\
Meta-pooling (proposed)  & $\bm{93.16 \pm 0.83}$ & $96.5 \pm 0.69$   \\
\hline
\end{tabular}
\end{table}

\begin{figure}[t]
  \centering
  \begin{minipage}[b]{0.45\linewidth}
    \centering
    \includegraphics[keepaspectratio, width=1.0\hsize]{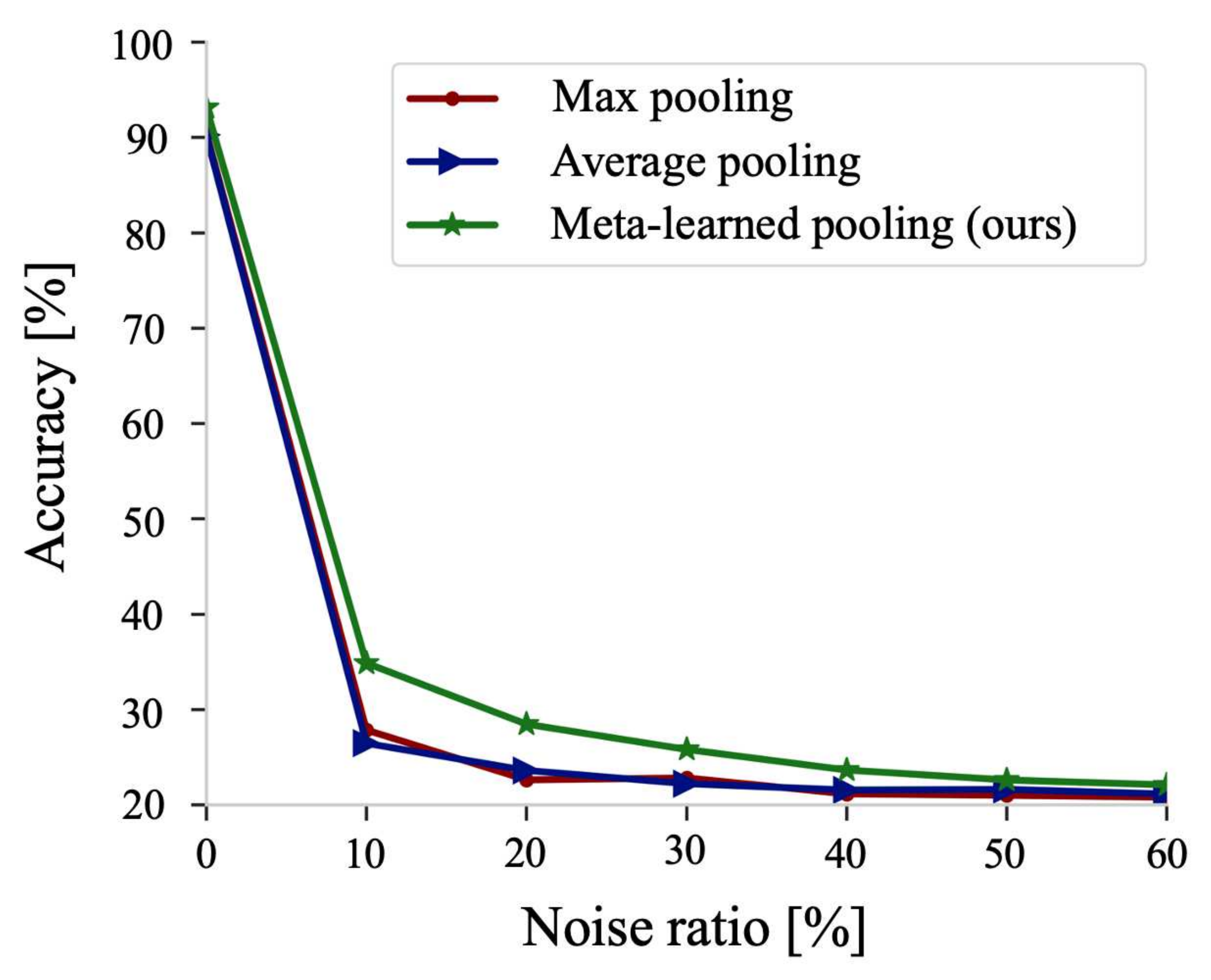}
    \subcaption{One-shot}\label{fig:acc1shot5way}
  \end{minipage}
  \begin{minipage}[b]{0.45\linewidth}
    \centering
    \includegraphics[keepaspectratio, width=0.95\hsize]{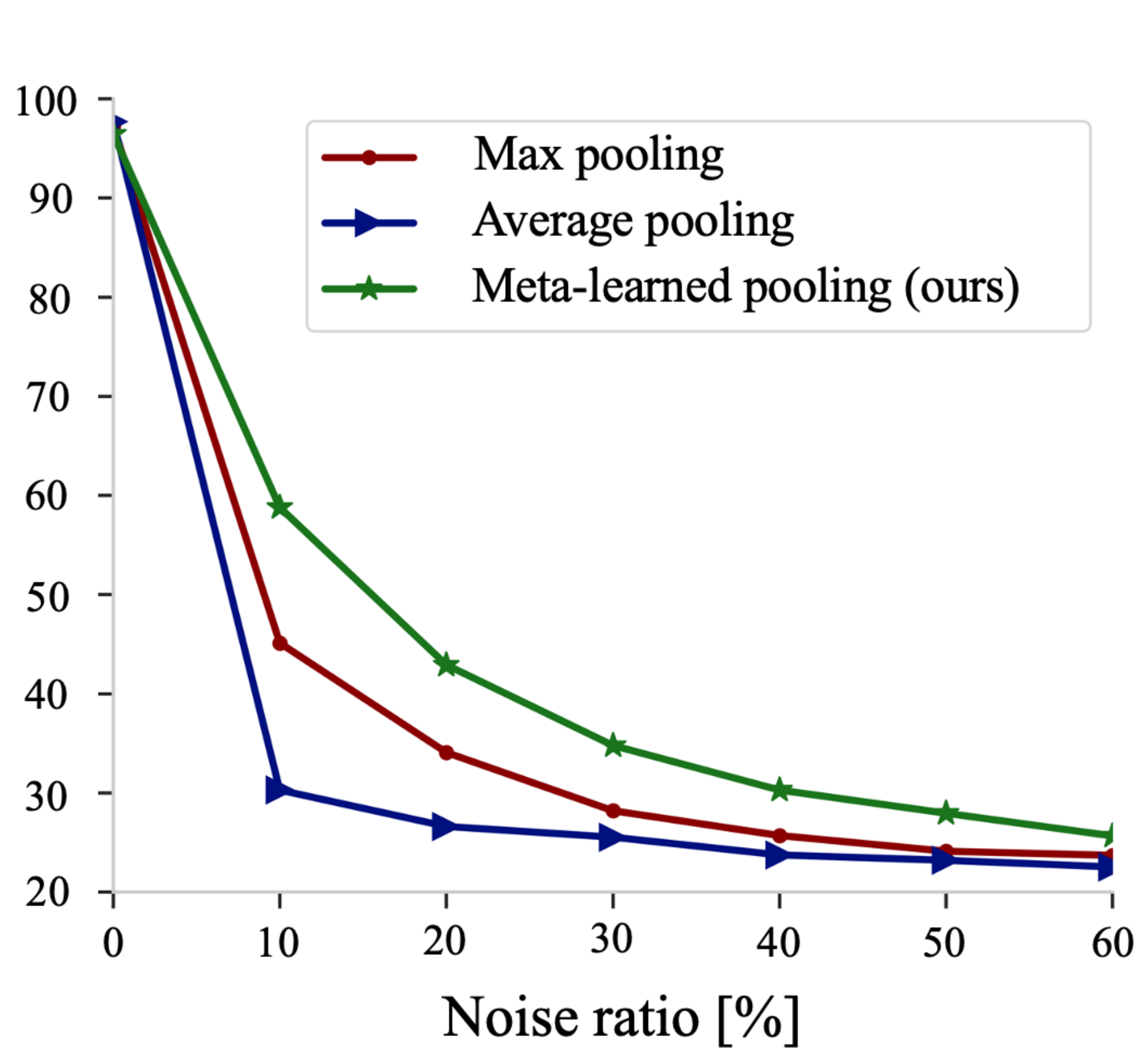}
    \subcaption{Five-shot}\label{fig:acc5shot5way}
  \end{minipage}
  \caption{Accuracy in the noisy image recognition task.}\label{fig:noise_result}
\end{figure}

\begin{figure}[!t]
    \centering
    \includegraphics[width=0.65\linewidth]{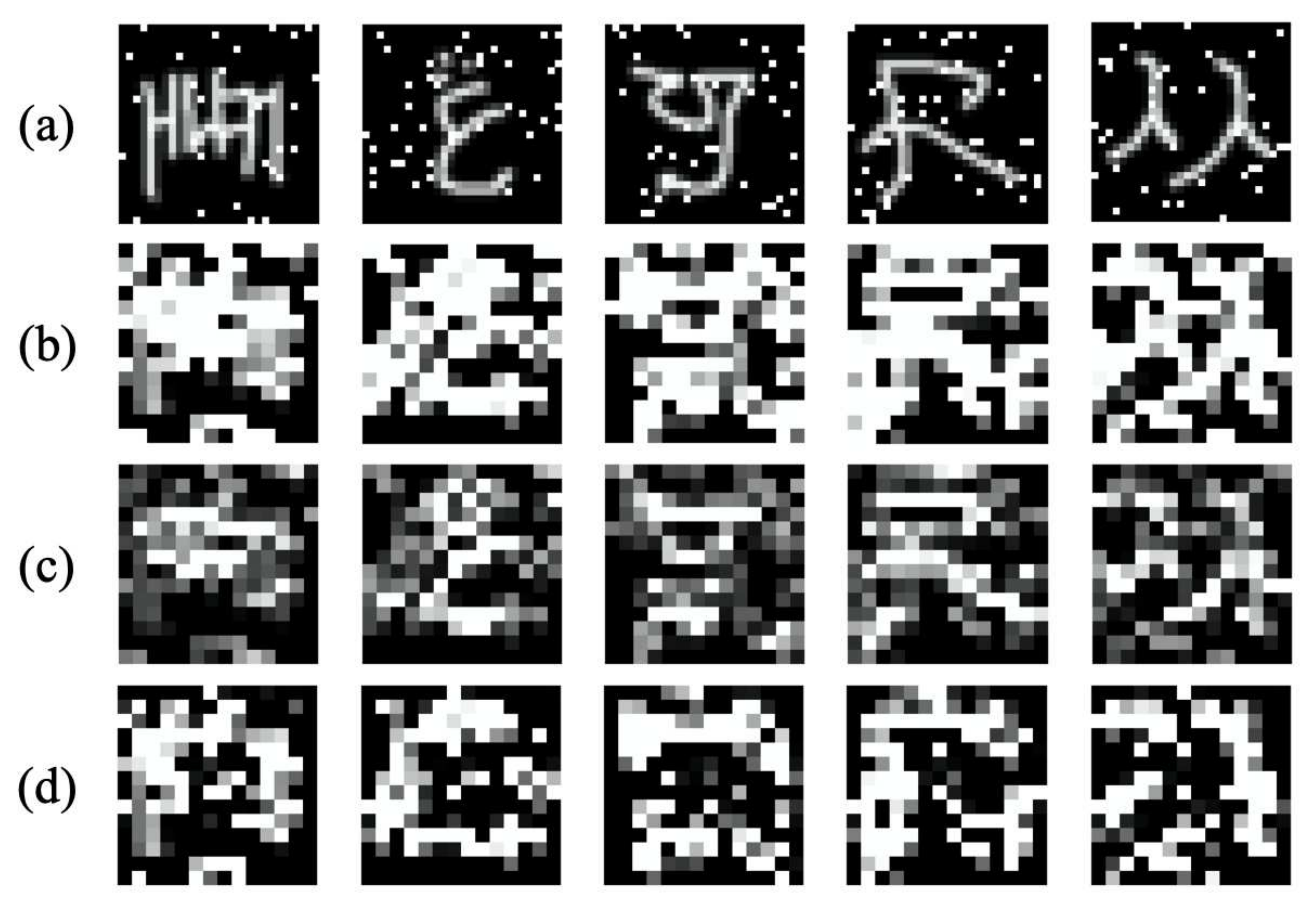}
    \caption{Pooling features for images with noise (noise ratio: 10\%): (a)~input image with noise, (b)~max pooling, (c)~average pooling, and (d)~meta-learned pooling.}
    \label{fig:noise-channel}
\end{figure}

The results of noisy image recognition are presented in Fig.~\ref{fig:noise_result}. The figures show the accuracy of each method with varying noise ratio. The proposed method outperformed max and average pooling when the noise ratio was large. Fig.~\ref{fig:noise-channel} exhibits the pooling features for each pooling operation when the noise ratio was 10\%. The figures display a randomly selected channel out of 64 channels for five examples of test images. It was confirmed that the max pooling features excessively emphasized noise, and the average pooling produced blurred features due to noise. In contrast, the proposed method could acquire relatively clean pooling features that preserved the strokes of the characters. These results suggest that the proposed method can learn a pooling layer with a high generalization capability, thus resulting in robustness to noise.\par

\section{Conclusion}\label{sec:conclusion}
In this paper, we proposed a meta-learning method for pooling layers. As part of our meta-learning framework, we proposed a parameterized pooling layer in which the kernel shape and pooling operation are trainable using two parameters, thereby allowing flexible pooling of the input data. We also proposed a meta-learning algorithm for the parameterized pooling layer, which allows us to acquire an appropriate pooling layer across the distribution of tasks. In the experiment, we applied the proposed meta-learning framework to character recognition tasks. The results demonstrated that a pooling layer that is suitable across multiple character recognition tasks was obtained via meta-learning, and the obtained pooling layer improved the performance of the model for new tasks of few-shot character image recognition and noisy character image recognition.\par

In future work, we will conduct a further comparison with other pooling methods since we only compared the proposed method with max pooling and average pooling. A detailed analysis of the computational complexity will also be performed.
\par\noindent\textbf{Acknowledgments:} This work was supported by JSPS KAKENHI Grant Numbers JP17K12752 and JP17H06100.

%
%

\bibliographystyle{splncs04}
\bibliography{icdar}

\begin{thebibliography}{10}
\providecommand{\url}[1]{\texttt{#1}}
\providecommand{\urlprefix}{URL }
\providecommand{\doi}[1]{https://doi.org/#1}

\bibitem{Baik_2020_CVPR}
Baik, S., Hong, S., Lee, K.M.: Learning to forget for meta-learning. In: IEEE
  Conference on Computer Vision and Pattern Recognition. pp. 2379--2387 (2020)

\bibitem{berman2019multigrain}
Berman, M., J{\'e}gou, H., Vedaldi, A., Kokkinos, I., Douze, M.: Multigrain: a
  unified image embedding for classes and instances. arXiv preprint
  arXiv:1902.05509  (2019)

\bibitem{chen2020variational}
Chen, J., Zhan, L.M., Wu, X.M., Chung, F.l.: Variational metric scaling for
  metric-based meta-learning. In: AAAI Conference on Artificial Intelligence.
  vol.~34, pp. 3478--3485 (2020)

\bibitem{Cui_2017_CVPR}
Cui, Y., Zhou, F., Wang, J., Liu, X., Lin, Y., Belongie, S.: Kernel pooling for
  convolutional neural networks. In: IEEE Conference on Computer Vision and
  Pattern Recognition. pp. 2921--2930 (2017)

\bibitem{dollar2009integral}
Dollar, P., Tu, Z., Perona, P., Belongie, S.: Integral channel features. In:
  British Machine Vision Conference (2009)

\bibitem{Elsken_2020_CVPR}
Elsken, T., Staffler, B., Metzen, J.H., Hutter, F.: Meta-learning of neural
  architectures for few-shot learning. In: IEEE Conference on Computer Vision
  and Pattern Recognition. pp. 12365--12375 (2020)

\bibitem{feng2011geometric}
Feng, J., Ni, B., Tian, Q., Yan, S.: Geometric $l_p$-norm feature pooling for
  image classification. In: IEEE Conference on Computer Vision and Pattern
  Recognition. pp. 2609--2704 (2011)

\bibitem{finn2017model}
Finn, C., Abbeel, P., Levine, S.: Model-agnostic meta-learning for fast
  adaptation of deep networks. In: International Conference on Machine
  Learning. pp. 1126--1135 (2017)

\bibitem{gao2016compact}
Gao, Y., Beijbom, O., Zhang, N., Darrell, T.: Compact bilinear pooling. In:
  IEEE Conference on Computer Vision and Pattern Recognition. pp. 317--326
  (2016)

\bibitem{gao2019global}
Gao, Z., Xie, J., Wang, Q., Li, P.: Global second-order pooling convolutional
  networks. In: IEEE Conference on Computer Vision and Pattern Recognition. pp.
  3024--3033 (2019)

\bibitem{Gao_2019_ICCV}
Gao, Z., Wang, L., Wu, G.: {LIP}: {Local} importance-based pooling. In:
  International Conference on Computer Vision. pp. 3355--3364 (2019)

\bibitem{graham2014fractional}
Graham, B.: Fractional max-pooling. arXiv preprint arXiv:1412.6071  (2014)

\bibitem{Hou_2020_CVPR}
Hou, Q., Zhang, L., Cheng, M.M., Feng, J.: Strip pooling: Rethinking spatial
  pooling for scene parsing. In: IEEE Conference on Computer Vision and Pattern
  Recognition. pp. 4003--4012 (2020)

\bibitem{ioffe2015batch}
Ioffe, S., Szegedy, C.: Batch normalization: Accelerating deep network training
  by reducing internal covariate shift. In: International Conference on Machine
  Learning. pp. 448--456 (2015)

\bibitem{khodadadeh2019unsupervised}
Khodadadeh, S., B{\"o}l{\"o}ni, L., Shah, M.: Unsupervised meta-learning for
  few-shot image classification. In: Advances in Neural Information Processing
  Systems. vol.~32 (2019)

\bibitem{kingma2014adam}
Kingma, D.P., Ba, J.: Adam: A method for stochastic optimization. In:
  International Conference on Learning Representations (2015)

\bibitem{lake2019omniglot}
Lake, B.M., Salakhutdinov, R., Tenenbaum, J.B.: The omniglot challenge: a
  3-year progress report. Current Opinion in Behavioral Sciences  \textbf{29},
  97--104 (2019)

\bibitem{li2018towards}
Li, P., Xie, J., Wang, Q., Gao, Z.: Towards faster training of global
  covariance pooling networks by iterative matrix square root normalization.
  In: IEEE Conference on Computer Vision and Pattern Recognition. pp. 947--955
  (2018)

\bibitem{lin2015bilinear}
Lin, T.Y., RoyChowdhury, A., Maji, S.: Bilinear {CNN} models for fine-grained
  visual recognition. In: IEEE International Conference on Computer Vision. pp.
  1449--1457 (2015)

\bibitem{Malinowski2013LearningSP}
Malinowski, M., Fritz, M.: Learning smooth pooling regions for visual
  recognition. In: British Machine Vision Conference (2013)

\bibitem{munkhdalai2017meta}
Munkhdalai, T., Yu, H.: Meta networks. In: International Conference on Machine
  Learning. pp. 2554--2563 (2017)

\bibitem{nguyenvan2019pooling}
NguyenVan, D., Lu, S., Tian, S., Ouarti, N., Mokhtari, M.: A pooling based
  scene text proposal technique for scene text reading in the wild. Pattern
  Recognition  \textbf{87},  118--129 (2019)

\bibitem{otsuzuki2020regularized}
Otsuzuki, T., Hayashi, H., Zheng, Y., Uchida, S.: Regularized pooling. In:
  International Conference on Artificial Neural Networks. pp. 241--254.
  Springer (2020)

\bibitem{rusu2018meta}
Rusu, A.A., Rao, D., Sygnowski, J., Vinyals, O., Pascanu, R., Osindero, S.,
  Hadsell, R.: Meta-learning with latent embedding optimization. In:
  International Conference on Learning Representations (2019)

\bibitem{saeedan2018detail}
Saeedan, F., Weber, N., Goesele, M., Roth, S.: Detail-preserving pooling in
  deep networks. In: IEEE Conference on Computer Vision and Pattern
  Recognition. pp. 9108--9116 (2018)

\bibitem{pmlr-v48-santoro16}
Santoro, A., Bartunov, S., Botvinick, M., Wierstra, D., Lillicrap, T.:
  Meta-learning with memory-augmented neural networks. In: International
  Conference on Machine Learning. vol.~48, pp. 1842--1850 (2016)

\bibitem{sermanet2012convolutional}
Sermanet, P., Chintala, S., LeCun, Y.: Convolutional neural networks applied to
  house numbers digit classification. In: International Conference on Pattern
  Recognition. pp. 3288--3291 (2012)

\bibitem{sung2018learning}
Sung, F., Yang, Y., Zhang, L., Xiang, T., Torr, P.H., Hospedales, T.M.:
  Learning to compare: Relation network for few-shot learning. In: IEEE
  Conference on Computer Vision and Pattern Recognition. pp. 1199--1208 (2018)

\bibitem{tolias2016particular}
Tolias, G., Sicre, R., J{\'e}gou, H.: Particular object retrieval with integral
  max-pooling of {CNN} activations. In: International Conference on Learning
  Representations (2016)

\bibitem{NIPS2016_90e13578}
Vinyals, O., Blundell, C., Lillicrap, T., kavukcuoglu, k., Wierstra, D.:
  Matching networks for one shot learning. In: Advances in Neural Information
  Processing Systems. vol.~29, pp. 3630--3638 (2016)

\bibitem{wang2018multi}
Wang, H., Wang, Q., Gao, M., Li, P., Zuo, W.: Multi-scale location-aware kernel
  representation for object detection. In: IEEE Conference on Computer Vision
  and Pattern Recognition. pp. 1248--1257 (2018)

\bibitem{Wei_2019_CVPR}
Wei, Z., Zhang, J., Liu, L., Zhu, F., Shen, F., Zhou, Y., Liu, S., Sun, Y.,
  Shao, L.: Building detail-sensitive semantic segmentation networks with
  polynomial pooling. In: IEEE Conference on Computer Vision and Pattern
  Recognition. pp. 7115--7123 (2019)

\bibitem{yu2014mixed}
Yu, D., Wang, H., Chen, P., Wei, Z.: Mixed pooling for convolutional neural
  networks. In: International Conference on Rough Sets and Knowledge
  Technology. pp. 364--375. Springer (2014)

\bibitem{zhao2017pyramid}
Zhao, H., Shi, J., Qi, X., Wang, X., Jia, J.: Pyramid scene parsing network.
  In: IEEE Conference on Computer Vision and Pattern Recognition. pp.
  2881--2890 (2017)

\bibitem{zhou2021metalearning}
Zhou, A., Knowles, T., Finn, C.: Meta-learning symmetries by
  reparameterization. In: International Conference on Learning Representations
  (2021)

\end{thebibliography}
%
\end{document}